\documentclass{article}
\usepackage{spconf,amsmath,graphicx,hyperref,booktabs}
\usepackage{kotex}

\title{UPP-Gauss: Ultrasound Propagation Physics-informed Gaussian Splatting for B-mode Novel-View Synthesis}

\name{\shortstack{
Guil Jung$^{1,*}$,\hspace{0.5cm}
Seok-Hwan Oh$^{1,*}$,\hspace{0.5cm}
Myeong-Gee Kim$^{1}$,\hspace{0.5cm}
Young-Min Kim$^{1}$,\hspace{0.5cm}
Hyeon-Jik Lee$^{2}$\\
Sang-Yun Kim$^{2}$,\hspace{0.5cm}
Hyeok-Jun Kwon$^{2}$,\hspace{0.5cm}
Hyuk-Sool Kwon$^{3}$,\hspace{0.5cm}
Hyeon-Min Bae$^{2}$
}}

\address{
$^{1}$Barreleye Inc., Seoul, South Korea\\
$^{2}$Department of Electrical Engineering, KAIST,
Daejeon, South Korea\\
$^{3}$Department of Emergency Medicine, SNUBH,
Seongnam, South Korea}

\begin{document}

\maketitle

\begingroup
\renewcommand{\thefootnote}{\fnsymbol{footnote}}
\footnotetext[1]{Guil Jung and Seok-Hwan Oh contributed equally to this work.}
\endgroup
\begin{abstract}
Accurate novel-view synthesis of B-mode ultrasound images requires modeling
acoustic propagation through tissue and the resulting appearance variations
across probe poses. We propose Ultrasound Propagation Physics-informed Gaussian
Splatting (UPP-Gauss), which integrates these acoustic interactions into an
explicit Gaussian representation. UPP-Gauss constructs acoustic maps by
sampling three 3D Gaussian fields using the probe pose and acoustic beam
propagation direction, then combines these maps according to ultrasound
propagation to synthesize B-mode images. Across three acoustically diverse
datasets, UPP-Gauss achieves the best reconstruction quality with efficient
rendering, improving PSNR over the best baseline on each dataset by
$1.43$~dB on average. These results demonstrate accurate and efficient B-mode
novel-view synthesis through explicit Gaussian modeling of ultrasound
propagation.
\end{abstract}

\begin{figure*}[t]
    \centering
    \includegraphics[width=\textwidth]{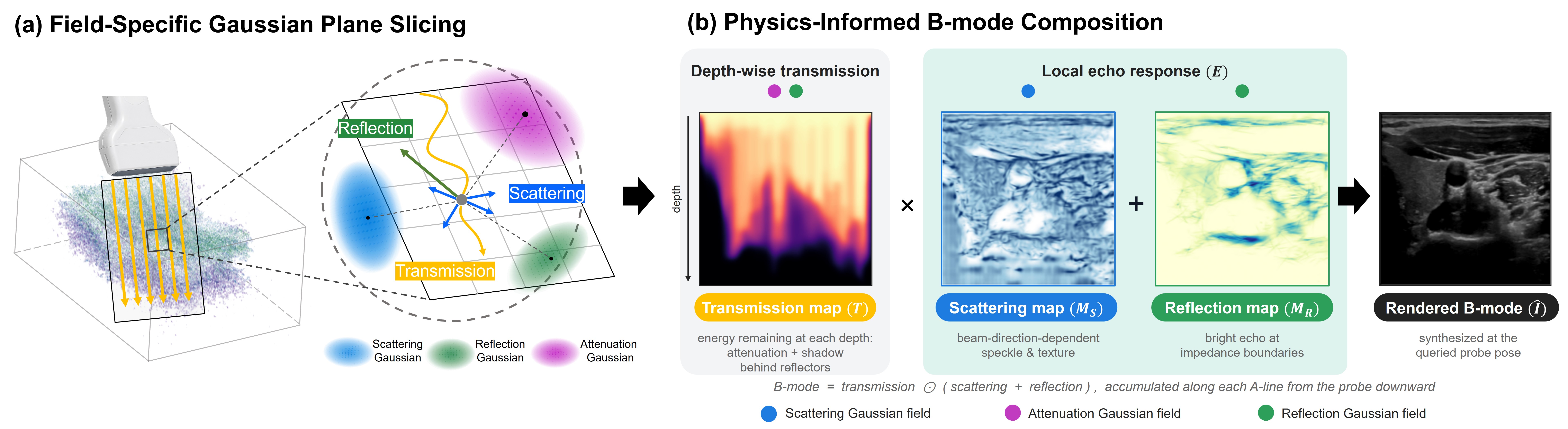}
    \caption{Overview of UPP-Gauss. The scattering, reflection, and attenuation
Gaussian fields are sampled on the probe imaging plane and combined through
local echo formation and depth-wise transmission to synthesize a B-mode image.}
    \label{fig:overall}
\end{figure*}

\section{Introduction}
\label{sec:introduction}

Ultrasound is widely used for clinical diagnosis as a real-time, noninvasive, and low-cost imaging modality. However, 2D B-mode imaging has a limited field of view, requiring clinicians to infer 3D anatomy from multiple
cross-sections~\cite{mozaffari2017freehand}. Ultrasound novel-view synthesis addresses this
limitation by integrating acquired frames into a consistent 3D representation
and synthesizing B-mode images at unseen probe poses. Neural Radiance Fields
(NeRF) and 3D Gaussian Splatting (3DGS) provide effective representations for
learning continuous 3D scenes from 2D observations
~\cite{mildenhall2021nerf,kerbl20233d}.

Unlike optical images formed from projected radiance, B-mode ultrasound arises
from waves propagating through tissue and returning through scattering and
reflection. Microscopic scatterers produce speckle and tissue texture, acoustic-impedance
boundaries generate bright reflections and posterior shadows, and attenuation
accumulates along propagation paths~\cite{jensen1991model,wagner1983statistics,kremkau1986artifacts}. The observed acoustic response of tissue microstructure at a given location can also vary with probe orientation and the preceding propagation path~\cite{szabo2004diagnostic,salehi2015patient}. Conventional optical projection models are therefore limited in their ability to capture these variations in B-mode appearance.

Prior learning-based ultrasound reconstruction has explored implicit neural representations. ImplicitVol~\cite{yeung2021implicitvol} reconstructs arbitrary imaging planes from a continuous 3D intensity field, while Ultra-NeRF~\cite{wysocki2024ultra} and UlRe-NeRF~\cite{guo2024ulre} introduce ultrasound image formation and direction-dependent reflection, respectively. These implicit representations require repeated network evaluations during training and rendering, incurring substantial computational costs. Recent methods use explicit Gaussian representations for efficient rendering. UltraGS~\cite{yang2025ultrags} models attenuation, reflection, and scattering as additive rendering terms, but does not explicitly formulate their interactions during acoustic propagation. UltraG-Ray~\cite{duelmer2026ultragray} models ray-wise energy decay using Gaussian echo intensity and transmittance. However, a unified B-mode formation model that captures beam-direction-dependent scattering and derives both local reflection echoes and downstream transmission loss from the reflection field remains underexplored.

We propose UPP-Gauss, which constructs scattering, reflection, and attenuation
maps by sampling Gaussian fields on the probe-defined imaging
plane. It combines local scattering and reflection echoes with depth-wise transmission
determined by accumulated attenuation and reflection losses to synthesize
B-mode images at queried poses. Our contributions are as follows. \textbf{(1)} We introduce an efficient plane-sliced Gaussian rendering scheme that evaluates acoustic Gaussian fields directly on the probe-defined imaging plane for physics-informed B-mode synthesis. \textbf{(2)} We model beam-direction-dependent
scattering and formulate each reflection response to account for both local
echo formation and downstream transmission loss. \textbf{(3)} Experiments on
three acoustically meaningful datasets demonstrate state-of-the-art reconstruction
quality, efficient rendering, and responses consistent with their distinct
acoustic characteristics. 

\section{Method}
\label{sec:method}

Fig.~\ref{fig:overall} illustrates the overall B-mode formation process of
UPP-Gauss. The scattering, reflection, and attenuation Gaussian fields are sampled on the
probe-defined imaging plane to construct the scattering map $M_S$, reflection
map $M_R$, and attenuation map $M_A$, respectively. Acoustic propagation is modeled along each A-line, with attenuation and reflection progressively reducing the transmitted intensity
with depth. We synthesize the B-mode image by modulating local echo
from $M_S$ and $M_R$ with depth-wise transmission accounting for attenuation
and reflection losses from $M_A$ and $M_R$.

\subsection{Plane-Sliced Gaussian Acoustic Fields}
\label{subsec:plane_sliced_fields}
In this section, we define the 2D imaging plane in world coordinates from the
probe pose and compute the spatial responses of the 3D Gaussian acoustic fields
on the imaging-plane grid.

For frame $n$, let $(\mathbf R_n,\mathbf t_n)$ denote the probe-to-world
transform, with $x$, $y$, and $z$ corresponding to the lateral, axial, and
elevational axes. Pixel $(u,v)$ has probe-coordinate position $(x_u,y_v,0)$. The corresponding
world-coordinate position is
\begin{equation}
\mathbf{p}_{nuv}
=
\mathbf{R}_n(x_u,y_v,0)^\top+\mathbf{t}_n .
\label{eq:world_sample}
\end{equation}

\noindent Each component $k\in\{S,R,A\}$, corresponding to scattering, reflection, and
attenuation, is represented by an independent anisotropic Gaussian set
$\mathcal G_k$ with centers $\boldsymbol{\mu}_g^k$ and covariances
$\boldsymbol{\Sigma}_g^k$ parameterized as in 3DGS~\cite{kerbl20233d}. 

For each Gaussian $g$, let
$\boldsymbol{\delta}_{ng}^k
=\mathbf{p}_{nuv}-\boldsymbol{\mu}_g^k$
denote the displacement from its center to the sampling point. We account for
its anisotropic covariance using the squared Mahalanobis distance
$(d_{ng}^k)^2=(\boldsymbol{\delta}_{ng}^k)^\top
(\boldsymbol{\Sigma}_g^k)^{-1}\boldsymbol{\delta}_{ng}^k$
and define the spatial weight as

\begin{equation}
w_{ng}^k(u,v)=
\begin{cases}
e^{-\frac{1}{2}(d_{ng}^k)^2},
& \|\boldsymbol{\delta}_{ng}^k\|_2\leq r,\\
0, & \text{otherwise}.
\end{cases}
\label{eq:local_gaussian_weight}
\end{equation}

\noindent We set $r=5\,\mathrm{mm}$ and evaluate only Gaussians centered within the
spherical support of each imaging-plane sample.

\subsection{Field-Specific Acoustic Map Construction}
\label{subsec:acoustic_maps}

For each component $k\in\{S,R,A\}$, the acoustic map is formed from the
weighted responses of its Gaussian field on the imaging plane:

\begin{equation}
M_{k,n}(u,v)
=
\sum_{g\in\mathcal{G}_k}
w_{ng}^{k}(u,v)
\left[a_g^{k}(\widehat{\mathbf d}_n)\right]_+,
\label{eq:acoustic_map}
\end{equation}

\noindent where $a_g^{k}(\widehat{\mathbf d}_n)$ is the directional response of Gaussian
$g$ for component $k$. The operator $[\cdot]_+$ denotes ReLU and constrains
each contribution to be non-negative.

In B-mode images, scattering appearance can vary with probe--tissue
orientation because tissue structure is observed through an anisotropic
point spread function (PSF) aligned with the acoustic beam~\cite{szabo2004diagnostic, thijssen1990texture}. Conventional 3DGS models view-dependent optical appearance using spherical harmonics (SH) evaluated along the viewing direction from the camera center to
each Gaussian~\cite{kerbl20233d}. Because B-mode images are formed along transducer A-lines rather than by pinhole-camera projection, we parameterize this directional response by the
acoustic beam direction. For a linear-array transducer, A-lines are modeled as parallel beams normal to its surface, with directional variations from Gaussian elevational offsets approximated by a shared axial direction:

\begin{equation}
\widehat{\mathbf d}_n
=
\mathbf R_n
\begin{bmatrix}
0 & 1 & 0
\end{bmatrix}^{\top},
\label{eq:beam_direction}
\end{equation}

\noindent where $[0,1,0]^\top$ is the axial direction in probe coordinates.
The Gaussian response along the direction is parameterized as

\begin{equation}
a_g^{k}(\widehat{\mathbf d}_n)
=
\sum_{\ell=0}^{D_k}
\sum_{m=-\ell}^{\ell}
c_{g,\ell m}^{k}
Y_{\ell m}(\widehat{\mathbf d}_n).
\label{eq:field_response}
\end{equation}

\noindent Where $Y_{\ell m}$ is an SH basis function, $c_{g,\ell m}^{k}$ is a learnable SH coefficient, and $D_k$ is the maximum SH degree.

\subsection{Physics-Informed B-Mode Composition}
\label{subsec:bmode_composition}

The observed B-mode intensity at each spatial location is modeled as the product of the local
echo response $E$ and depth-wise transmission $T$, which accounts for
cumulative loss along the preceding propagation path
~\cite{salehi2015patient,wysocki2024ultra}. Omitting the frame index $n$, we
define the final B-mode
intensity as
\begin{equation}
\widehat{I}(u,v)=T(u,v)E(u,v).
\label{eq:bmode_intensity}
\end{equation}

The local echo response is the sum of the scattering and reflection
contributions:
\begin{equation}
E(u,v)=M_S(u,v)+\overline{M}_R(u,v).
\label{eq:local_echo}
\end{equation}

\noindent We transform $M_R$ into the local reflection response
$\overline{M}_R$:
\begin{equation}
\overline{M}_R(u,v)=1-\exp[-M_R(u,v)\Delta s].
\label{eq:bounded_reflection}
\end{equation}

\noindent For uniformly spaced axial rows, $\Delta s=y_{v+1}-y_v$ is the
physical depth spacing and path-length increment along an A-line. Since
$M_R\geq0$ by the ReLU in Eq.~\eqref{eq:acoustic_map} and $\Delta s>0$,
Eq.~\eqref{eq:bounded_reflection} yields a bounded local reflection response.

Each reflection Gaussian learns a shadow-coupling weight $\gamma_g\in[0,1]$.
We define $M_{R,\gamma}=\sum_{g\in\mathcal G_R}\gamma_g w_g^R[a_g^R]_+$
and $\overline{M}_{R,\gamma}=1-\exp(-M_{R,\gamma}\Delta s)$. The factor $1-\overline{M}_{R,\gamma}$
represents transmission after weighted reflection loss and is accumulated
with attenuation over preceding axial samples:
{\small
\begin{equation}
\begin{aligned}
T(u,v)
&=I_0\prod_{v'<v}
\exp[-M_A(u,v')\Delta s]
[1-\overline{M}_{R,\gamma}(u,v')]\\[-1mm]
&=I_0\exp\!\left[
-\Delta s\sum_{v'<v}
\left\{M_A(u,v')+M_{R,\gamma}(u,v')\right\}
\right].
\end{aligned}
\label{eq:transmission}
\end{equation}
}

\noindent Eq.~\eqref{eq:transmission} accumulates attenuation and weighted
reflection factors, reducing their product to an exponential sum. The magnitude of reflection-induced transmission loss varies across tissue interfaces. To account for this variability, we assigned a learnable weight $\gamma_g$ for each reflection Gaussian to parameterize its contribution to transmission loss. Accumulation over $v'<v$ applies reflection loss only at subsequent depths, preserving the local echo at the current location.
$I_0$ denotes the initial incident intensity and is normalized to $I_0=1$
in all experiments.

\subsection{Optimization}
\label{subsec:optimization}

The Gaussian parameters are jointly optimized using the photometric loss
\begin{equation}
\mathcal{L}_{\mathrm{obj}}
=
0.9\!\left(1-\operatorname{SSIM}
(\widehat{I},I^{\mathrm{gt}})\right)
+
0.1\operatorname{MSE}
(\widehat{I},I^{\mathrm{gt}}).
\label{eq:photometric_loss}
\end{equation}

\noindent Each Gaussian field is initialized on a regular $50^3$ grid within the
world-space bounding box of the training image corners, from which Gaussians
outside the training imaging planes expanded by a $3$~mm elevational margin
are removed before optimization.

\begin{figure}[t]
\centering
\includegraphics[width=\columnwidth]{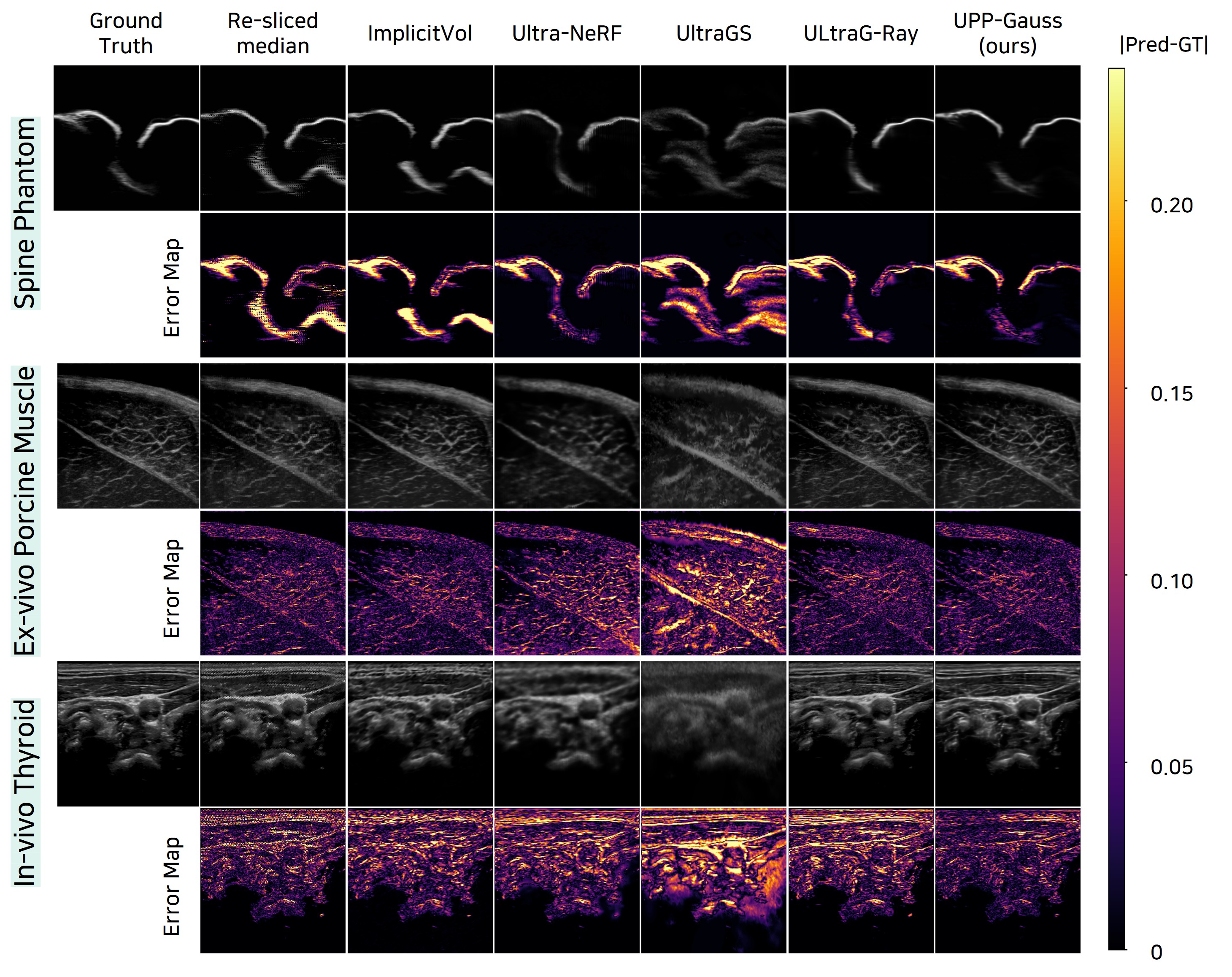}
\caption{Qualitative comparison on spine, porcine muscle, and thyroid. Error maps show
absolute errors.}
\label{fig:comparison}
\end{figure}

\section{Experiments}
\label{sec:experiments}

\subsection{Experimental Setup}
\label{subsec:experimental_setup}

\noindent\textbf{Dataset.}
We evaluated UPP-Gauss on the publicly available spine phantom and ex-vivo porcine muscle datasets ~\cite{duelmer2026ultragray}, following the official data split. To evaluate UPP-Gauss on in-vivo data acquired in a clinical setting, we additionally acquired transverse freehand thyroid sweep from each of 13 participants using a GE Healthcare LOGIQ P9 system with an L3-12 linear probe. Probe poses were tracked using a Polaris Vega XT. The thyroid field of view is $5.1\times4.0$~cm
(width $\times$ depth). Each thyroid sweep contains 120 frames, with the
central 15 consecutive frames used for testing and the remainder for training.
Angular coverage is $40^\circ$, $6^\circ$, and an average of $8^\circ$ for
spine phantom, porcine muscle, and thyroid, respectively. The datasets
represent strong reflections and acoustic shadows, soft-tissue scattering,
and heterogeneous in-vivo anatomy, respectively.

\noindent\textbf{Evaluation Metrics and Baselines.}
We evaluate reconstruction quality using PSNR and SSIM~\cite{wang2004image}, and use the gradient-based GMS and GMSD metrics~\cite{xue2013gradient} to measure gradient-based perceptual similarity. We compare
against the conventional Re-sliced median Compounding method
~\cite{duelmer2026ultragray} and recent learning-based methods, ImplicitVol~\cite{yeung2021implicitvol}, Ultra-NeRF
~\cite{wysocki2024ultra}, UltraGS~\cite{yang2025ultrags}, and UltraG-Ray
~\cite{duelmer2026ultragray}, with public implementations.

\noindent\textbf{Implementation Details.}
All images are resized to $224\times224$. UPP-Gauss is trained for 50k iterations using Adam with a learning rate of $10^{-3}$. To model scattering responses that vary with probe orientation while maintaining computational efficiency, we set the SH degrees to $(D_S,D_R,D_A)=(2,0,0)$. Computational cost is measured on the spine phantom dataset using an NVIDIA RTX 3090 GPU.

\subsection{Comparative study}
\label{subsec:comparison}

Table~\ref{tab:comparison} and Fig.~\ref{fig:comparison} report quantitative
and qualitative results, respectively, with complexity and rendering costs in
Table~\ref{tab:computational_cost}.

UPP-Gauss performs best across all datasets and metrics. Compared with the best baseline for each metric, it improves
PSNR/SSIM by $2.13$~dB/$0.0277$ on spine phantom and
$1.51$~dB/$0.0749$ on thyroid, both characterized by strong boundary
reflections and propagation attenuation. As shown in
Fig.~\ref{fig:comparison}, UPP-Gauss reproduces bright reflection boundaries,
posterior acoustic shadows, and depth-dependent echo decay in spine phantom.
On porcine muscle and thyroid, it reconstructs scattering texture and boundary
echoes with lower errors than the compared methods.

UPP-Gauss renders a frame in $1.49$~ms with $0.33$~GFLOPs.  This demonstrates
that the proposed ultrasound renderer enables accurate and efficient B-mode synthesis
by directly sampling explicit Gaussian fields on the imaging plane.

\begin{table}[t]
\centering
\caption{Quantitative reconstruction comparison. Results are reported as
mean $\pm$ standard deviation across test frames.}
\label{tab:comparison}
\setlength{\tabcolsep}{3pt}
\renewcommand{\arraystretch}{0.95}
\resizebox{\columnwidth}{!}{%
\begin{tabular}{lcccc}
\toprule
Method & PSNR$\uparrow$ & SSIM$\uparrow$
       & GMS$\uparrow$ & GMSD$\downarrow$ \\
\midrule
\multicolumn{5}{c}{Spine Phantom} \\
\midrule
Re-sliced Median
& $25.18\pm2.97$ & $0.8959\pm0.0557$
& $0.9512\pm0.0276$ & $0.1429\pm0.0513$ \\
ImplicitVol
& $23.65\pm3.13$ & $0.9134\pm0.0386$
& $0.9484\pm0.0262$ & $0.1665\pm0.0538$ \\
Ultra-NeRF
& $25.80\pm0.86$ & $0.8419\pm0.0305$
& $0.9640\pm0.0104$ & $0.1154\pm0.0182$ \\
UltraGS
& $22.90\pm1.91$ & $0.7371\pm0.0607$
& $0.9182\pm0.0334$ & $0.1787\pm0.0408$ \\
UltraG-Ray
& $27.34\pm2.13$ & $0.8316\pm0.0285$
& $0.9669\pm0.0152$ & $0.1105\pm0.0327$ \\
UPP-Gauss (Ours)
& $\mathbf{29.47}\pm2.94$ & $\mathbf{0.9411}\pm0.0345$
& $\mathbf{0.9760}\pm0.0151$ & $\mathbf{0.0828}\pm0.0349$ \\
\midrule
\multicolumn{5}{c}{Ex-vivo Porcine Muscle} \\
\midrule
Re-sliced Median
& $25.83\pm1.30$ & $0.6673\pm0.0418$
& $0.9110\pm0.0156$ & $0.1317\pm0.0209$ \\
ImplicitVol
& $26.35\pm0.65$ & $0.6933\pm0.0257$
& $0.9174\pm0.0077$ & $0.1229\pm0.0085$ \\
Ultra-NeRF
& $24.40\pm0.61$ & $0.5784\pm0.0388$
& $0.8854\pm0.0126$ & $0.1546\pm0.0106$ \\
UltraGS
& $23.38\pm0.52$ & $0.4878\pm0.0204$
& $0.8104\pm0.0132$ & $0.2111\pm0.0077$ \\
UltraG-Ray
& $25.89\pm1.03$ & $0.6656\pm0.0495$
& $0.9143\pm0.0133$ & $0.1265\pm0.0141$ \\
UPP-Gauss (Ours)
& $\mathbf{26.99}\pm0.67$ & $\mathbf{0.7306}\pm0.0254$
& $\mathbf{0.9263}\pm0.0077$ & $\mathbf{0.1127}\pm0.0081$ \\
\midrule
\multicolumn{5}{c}{In-vivo Thyroid} \\
\midrule
Re-sliced Median
& $21.95\pm1.25$ & $0.5649\pm0.0617$
& $0.8412\pm0.0353$ & $0.2154\pm0.0322$ \\
ImplicitVol
& $22.08\pm1.52$ & $0.6055\pm0.0723$
& $0.8837\pm0.0230$ & $0.1746\pm0.0410$ \\
Ultra-NeRF
& $21.27\pm2.30$ & $0.5456\pm0.0850$
& $0.8475\pm0.0460$ & $0.2110\pm0.0415$ \\
UltraGS
& $20.80\pm1.14$ & $0.4995\pm0.0534$
& $0.8343\pm0.0210$ & $0.2226\pm0.0195$ \\
UltraG-Ray
& $22.56\pm1.31$ & $0.6076\pm0.0669$
& $0.8882\pm0.0180$ & $0.1714\pm0.0166$ \\
UPP-Gauss (Ours)
& $\mathbf{24.07}\pm1.58$ & $\mathbf{0.6825}\pm0.0680$
& $\mathbf{0.9061}\pm0.0201$ & $\mathbf{0.1475}\pm0.0215$ \\
\bottomrule
\end{tabular}%
}
\end{table}

\begin{table}[t]
\centering
\caption{Comparison of model complexity and rendering cost.}
\label{tab:computational_cost}

\small
\setlength{\tabcolsep}{3pt}
\renewcommand{\arraystretch}{0.82}

\begin{tabular*}{\columnwidth}{
    @{\extracolsep{\fill}} l c c c @{}
}
\toprule
Method
& Params (M)
& Latency (ms/frame)
& GFLOPs \\
\midrule
Re-sliced Median & --             & 0.54 & --             \\
ImplicitVol       & 0.157          & 4.61          & 15.70          \\
Ultra-NeRF        & 0.132 & 8.98          & 13.20          \\
UltraGS           & 3.916          & 2.42          & 0.67           \\
UltraG-Ray        & 1.767          & 2.50          & 0.91           \\
UPP-Gauss         & 4.160          & 1.49          & 0.33 \\
\bottomrule
\end{tabular*}
\end{table}

\subsection{Ablation Studies}
\label{subsec:ablation}
\begin{figure}[t]
\centering
\includegraphics[width=\columnwidth]{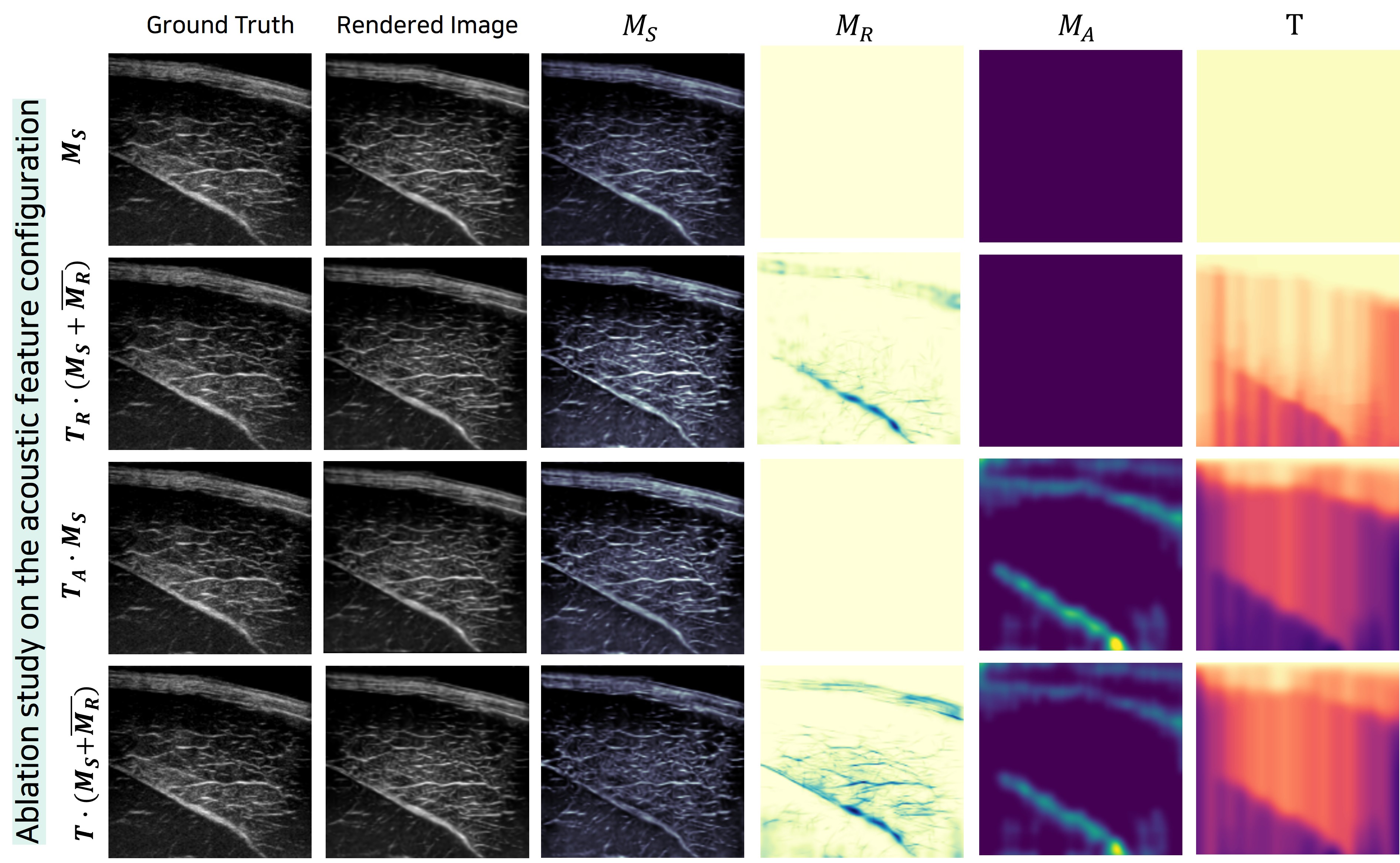}
\caption{Qualitative acoustic-map ablation on porcine muscle.}
\label{fig:map_ablation}
\end{figure}

To assess the contribution of the modeled propagation mechanisms, we separately train and evaluate
variants with $M_R$, $M_A$, or both ablated. Table~\ref{tab:map_ablation} and Fig.~\ref{fig:map_ablation} present the quantitative and qualitative results, respectively. $T_A$ and $T_R$ are the depth-wise transmission
factors derived from $M_A$ and $M_R$, respectively, with $T=T_AT_R$ for
$I_0=1$.

In the $M_S$-only configuration, $M_S$ absorbs tissue structure, reflection
boundaries, and depth-dependent intensity variation, reducing average PSNR by
$0.91$~dB across the three datasets. Adding $M_R$ assigns boundary echoes to $\overline{M}_R$ and downstream shadows to $T_R$, while $M_A$ introduces cumulative A-line attenuation through $T_A$. In the full configuration,
$M_R$ captures strong reflectors and fine internal boundaries, allowing
$M_S$ to focus on scattering structure and texture. This configuration outperforms all ablations across three datasets, supporting the effectiveness of our physics-informed formulation for ultrasound novel-view synthesis.

To assess the proposed beam-dependent scattering model, we evaluate
$D_S\in\{0,1,2\}$. At $D_S=1,2$, PSNR improves over $D_S=0$ by
$2.11/2.20$~dB on spine phantom, $0.18/0.26$~dB on porcine muscle, and
$0.49/0.75$~dB on thyroid. These results demonstrate the benefit of modeling scattering variations with
probe direction.

\begin{table}[t]
\centering
\caption{Ablation study of acoustic map combinations. Results are reported as
PSNR/SSIM.}
\label{tab:map_ablation}
\setlength{\tabcolsep}{3pt}
\renewcommand{\arraystretch}{1}
\resizebox{\columnwidth}{!}{%
\begin{tabular}{lccc}
\toprule
Composition & Spine phantom & Porcine muscle & Thyroid \\
\midrule
$M_S$
& $28.33/0.8482$ & $26.63/0.7115$ & $22.85/0.6285$ \\
$T_R(M_S+\overline{M}_R)$
& $28.33/0.8552$ & $26.64/0.7173$ & $22.88/0.6620$ \\
$T_A M_S$
& $29.00/0.9234$ & $26.67/0.7120$ & $23.58/0.6569$ \\
$T_A T_R(M_S+\overline{M}_R)$
& $\mathbf{29.47/0.9411}$
& $\mathbf{26.99/0.7306}$
& $\mathbf{24.07/0.6825}$ \\
\bottomrule
\end{tabular}%
}
\end{table}

\section{Conclusion}
\label{sec:conclusion}
We presented UPP-Gauss, a Gaussian representation for B-mode novel-view
synthesis that models scattering, reflection, and attenuation during ultrasound
propagation. It achieves improved reconstruction quality and efficient
rendering across acoustically diverse datasets, while the learned maps capture
tissue scattering, reflection boundaries, and depth-wise attenuation. Future work will extend and evaluate UPP-Gauss for diverse probe geometries in addition to linear arrays.

\vfill\pagebreak

\section{Compliance with Ethical Standards}
\label{sec:compliance}
The in-vivo thyroid data acquired for this study were collected under a
protocol approved by the relevant institutional review board (IRB).
OpenAI Codex was used for language editing of the manuscript.

\section{Acknowledgments}
The authors declare no conflicts of interest.

\bibliographystyle{IEEEbib}
\bibliography{strings}

\end{document}